%% file: paper.tex
\documentclass[conference]{IEEEtran}
\IEEEoverridecommandlockouts
\usepackage{cite}
\usepackage{amsmath,amssymb,amsfonts}
\usepackage{algorithmic}
\usepackage{graphicx}
\usepackage{textcomp}
\usepackage[dvipsnames]{xcolor} 
\usepackage{url} 
\usepackage{enumitem} 
\usepackage{color} 
\usepackage{soul} 
\usepackage{stfloats}
\usepackage{tikz} 
\usetikzlibrary{shapes.geometric}
\usetikzlibrary{arrows}
\usetikzlibrary{calc}
\usetikzlibrary{decorations.pathreplacing}
\usetikzlibrary{shadows}
\usetikzlibrary{positioning}
\usepackage{varwidth}
\tikzset{
  invisible/.style={opacity=0},
  visible on/.style={alt={#1{}{invisible}}},
  alt/.code args={<#1>#2#3}{%
    \alt<#1>{\pgfkeysalso{#2}}{\pgfkeysalso{#3}} 
  },
}

\tikzset{main node/.style={draw,rounded corners},
} 

\def\BibTeX{{\rm B\kern-.05em{\sc i\kern-.025em b}\kern-.08em
    T\kern-.1667em\lower.7ex\hbox{E}\kern-.125emX}}
\begin{document}

\input{macros}

\title{SoK: Taming the Triangle - On the Interplays between Fairness, Interpretability and Privacy in Machine Learning} 

\input{authors}

\maketitle 

\begin{abstract}
Machine learning techniques are increasingly used for high-stakes decision-making, such as college admissions, loan attribution or recidivism prediction.
Thus, it is  crucial to ensure that the models learnt can be audited or understood by human users, do not create or reproduce discrimination or bias, and do not leak sensitive information regarding their training data.
Indeed, interpretability, fairness and privacy are key requirements for the development of responsible machine learning, and all three have been studied extensively during the last decade.
However, they were mainly considered in isolation, while in practice they interplay with each other, either positively or negatively.
In this Systematization of Knowledge (SoK) paper, we survey the literature on the interactions between these three desiderata.
More precisely, for each pairwise interaction, we summarize the identified synergies and tensions.
These findings highlight several fundamental theoretical and empirical conflicts, while also demonstrating that jointly considering these different requirements is challenging when one aims at preserving a high level of utility.
To solve this issue, we also discuss possible conciliation mechanisms, showing that a careful design can enable to successfully handle these different concerns in practice.
\end{abstract}


\section{Introduction}\label{sec:introduction}
\input{0-introduction}

\section{Background}\label{sec:background} 
\input{1-background}

\section{Fairness and Interpretability}\label{sec:fairness_interpretability} 
\input{2-fairness-interpretability}

\section{Fairness and Privacy}\label{sec:fairness_privacy} 
\input{3-fairness_privacy}

\section{Interpretability and Privacy}\label{sec:interpretability_privacy} 
\input{4-interpretability_privacy}

\section{Conclusion}\label{sec:conclusion} 
\input{5-conclusion}


\bibliographystyle{IEEEtran}
\bibliography{IEEEabrv,references}


\appendices

\input{999-appendix_summary}

\end{document}

%% file: macros.tex
\def\learningalgo{\mathcal{L}}
\def\classifier{h}
\def\dataset{\mathcal{D}}
\def\nbexamples{N}

\def\dpsgd{\texttt{DP-SGD}}
\def\pate{\texttt{PATE}}
\def\pflr{\texttt{PFLR}}
\def\pflrs{\texttt{PFLR\textsuperscript{*}}}

\def\mjo#1{\textcolor{cyan}{#1}}
\def\rqe#1{\textcolor{orange}{#1}}

\def\revision#1{\textcolor{black}{#1}}
\def\revisionmultiline{\color{black}}

%% file: authors.tex
\author{\IEEEauthorblockN{1\textsuperscript{st} Julien Ferry}
\IEEEauthorblockA{
\textit{LAAS-CNRS, Universit\'{e} de Toulouse, CNRS}\\
Toulouse, France \\
jferry@laas.fr}
\and
\IEEEauthorblockN{2\textsuperscript{nd} Ulrich A{\"i}vodji}
\IEEEauthorblockA{\textit{\'Ecole de Technologie Sup\'erieure} \\
Montr\'eal, Canada \\
Ulrich.Aivodji@etsmtl.ca}
\and
\IEEEauthorblockN{3\textsuperscript{rd} S\'ebastien Gambs}
\IEEEauthorblockA{\textit{Universit\'e du Qu\'ebec \`a Montr\'eal} \\
Montr\'eal, Canada \\
gambs.sebastien@uqam.ca}
\and
\IEEEauthorblockN{4\textsuperscript{th} Marie-Jos\'e Huguet}
\IEEEauthorblockA{
\textit{LAAS-CNRS, Universit\'{e} de Toulouse, CNRS, INSA}\\
Toulouse, France \\
huguet@laas.fr}
\and
\IEEEauthorblockN{5\textsuperscript{th} Mohamed Siala}
\IEEEauthorblockA{
\textit{LAAS-CNRS, Universit\'{e} de Toulouse, CNRS, INSA}\\
Toulouse, France \\
msiala@laas.fr}
}

%% file: 0-introduction.tex
Machine learning (ML) models have many useful and promising applications. 
For instance, they can help to analyze medical data, which is becoming increasingly complex due to the improvements in medical tools. 
However, their growing use for high-stakes decision-making systems - such as college admissions, recidivism prediction or credit scoring - raises significant ethical, philosophical and societal challenges.
This has led to the regulation of their use through several legal texts, such as the European Union General Data Protection Regulation\footnote{\url{https://gdpr-info.eu/}}~\cite{voigt2017eu} or the forthcoming AI Act\footnote{\url{https://artificialintelligenceact.eu/}}.

In particular, three important ethical issues have emerged, each corresponding to a key concern that should be addressed to both comply with these new legal frameworks and lay the foundations towards a responsible ML.
First, ML algorithms require large amounts of data, which often contains personal information. 
Thus, it is of paramount importance to ensure that the \emph{privacy} of the involved individuals is not harmed while also being able to extract useful generic patterns from this data.
Second, it was shown that data-driven decision-making processes can create or reproduce biases that systematically disadvantage specific individuals or groups~\cite{DBLP:journals/csur/MehrabiMSLG21}. 
Quantifying but also reducing/eliminating these biases to promote \emph{fairness} is hence an important challenge.
Third, while common ML models, such as deep neural networks, can reach high predictive performance, their underlying logic and representation are often too complex, preventing users from fully understanding their decisions. 
This raises significant concerns, regarding their auditability, certifiability and trust, thus calling for the requirement of \emph{interpretability} with respect to their predictions.

These three topics, namely privacy, fairness and explainability, have been extensively studied during the last decade~\cite{DBLP:journals/corr/abs-2005-08679,barocas-hardt-narayanan,guidotti2018survey} with an emphasis on how they each trade-off with utility.
However, they are often considered in isolation, while in practice it is necessary to enforce them \emph{simultaneously}.
Characterizing their mutual interplays is hence an important research avenue, which has attracted some attention in the last years. 
Indeed, these concerns often conflict~\cite{datta2023position}, and trade-offs between them, as well as with utility, generally have to be set.
Throughout this SoK paper, we conduct an in-depth survey of the literature on the different compatibilities, synergies and tensions that have been identified between them.
\revision{More precisely, we focus on the supervised learning setup, and consider mainly classification tasks.}

\clearpage 

\paragraph*{\textbf{Positioning with respect to other surveys}}
Other recent works survey the literature on the interactions between 
several of our three identified desiderata. 
\revision{Among others, \cite{datta2023position} reviews at a high level the main tensions that occur between the human values of privacy, transparency and fairness when they have to be embodied in a machine learning model. We extend this work by additionally considering compatibilities and synergies. Furthermore, we focus on the interplays between the three aspects to allow a more thorough technical discussion.}
Furthermore,~\cite{DBLP:conf/ijcai/Fioretto0HZ22}, investigates solely the interplays between fairness and (differential) privacy by conducting an in-depth analysis on how one influences the other.
\revision{We extend this study in Section~\ref{sec:fairness_privacy}.}
Finally, a recent thesis~\cite{schoffer2023interplay} focuses on the interactions between transparency and fairness. It provides a deepening of (part of) our dedicated Section~\ref{sec:fairness_interpretability}.

The outline of the paper is as follows. 
First in Section~\ref{sec:background}, we review the background regarding the three considered aspects of responsible ML, namely fairness, interpretability and privacy before surveying their interplays.
More precisely, Section~\ref{sec:fairness_interpretability} considers both fairness and interpretability, Section~\ref{sec:fairness_privacy} studies the interactions between fairness and privacy, and Section~\ref{sec:interpretability_privacy} summarizes the connections between interpretability and privacy.
We then conclude with the identified key challenges in 
Section~\ref{sec:conclusion}.
Finally, Appendix~\ref{appendix:summary} provides a graphical summary of all the analyzed interplays.

%% file: 1-background.tex
In this section, we introduce the three identified pillars of responsible machine learning. 
For each of them, we briefly review their key ideas, with an emphasize on the particular aspects that will ease the understanding of subsequent sections.

\subsection{Fairness}\label{subsec:fairness}

Different approaches to fairness have been proposed in the literature, which can be grouped into three main categories~\cite{DBLP:conf/icse/VermaR18}. 
The rationale of \emph{statistical fairness}, also coined \emph{group fairness}, is to ensure that a given statistical measure has similar values between several \emph{subgroups}, defined by the value(s) of some sensitive feature(s).
For example, the statistical parity fairness metric aims at equalizing the positive prediction rate across the different groups, while the equal opportunity metric considers the groups' true positive rates and finally the equalized odds metric handles both their true positive and true negative rates.
The underlying principle is that such sensitive features (\emph{e.g.}, race, gender, \ldots) should not influence the predictions. 
\emph{Individual fairness} approaches build on the idea that similar individuals should be treated similarly~\cite{Dwork2011}.
For instance, this can be formulated as a Lipschitz condition over the classification function, in which bounding the distance between two examples also bound the distance between their outputs from the model.
\emph{Causal fairness} techniques analyze the causal relationships between sensitive features, non-sensitive ones and the target decision, leveraging causal graphs~\cite{DBLP:conf/nips/KilbertusRPHJS17}. 

Depending on which step of the (supervised) ML pipeline they intervene on, fairness-enhancing methods can be divided into three main categories~\cite{DBLP:journals/ibmrd/BellamyDHHHKLMM19,DBLP:conf/fat/FriedlerSVCHR19,caton2020fairness}.
\emph{Pre-processing} methods aim at removing undesired correlations from the training data before applying standard learning techniques on the sanitized data while \emph{post-processing} techniques modify the outputs of a trained model to achieve fairness.
Finally, \emph{in-processing} (also called \emph{algorithmic modification}) techniques directly adapt the learning procedure to produce inherently fair models. 

\subsection{Explainability/Interpretability}
\label{subsec:interpretability}

There are two main approaches towards facilitating the understanding of ML models. 
On the one hand, post-hoc explanations~\cite{guidotti2018survey} can be crafted to explain the behaviour of a black-box model. 
Depending on their form, different types of such explanations can be defined, among which \emph{example-based explanations} consist in datapoints, belonging to the same space as the model's training set examples. 
For instance, they can be highly influential training examples~\cite{koh2017understanding}, nearest neighbours or prototypes. 
\emph{Counterfactual explanations} also fall into this category, as they are datapoints close to the explained instance but exhibiting a different prediction from the considered model.
\emph{Feature-based explanations} take the form of a vector in the feature space, in which each coordinate is the degree to which the associated feature influences a model's prediction. For example, in computer vision, saliency maps~\cite{selvaraju2017grad} highlight the regions of an input image that most contributed to the model's decision.
Feature-based explanations can be computed using several mechanisms. 
For instance, \emph{gradient-based} methods compute the gradients of a model (\emph{e.g.}, a deep neural network) with respect to the input features, either for a given class or for intermediate component(s) of the network, which enables to determine which features contribute the most to a particular prediction. 
In constrast, \emph{perturbation-based} methods modify the input provided to the black-box and observe the resulting changes in the model's outputs. 

On the other hand, one can learn models that are inherently interpretable by humans. 
For instance, decision trees or rule lists of reasonable size are commonly considered as interpretable~\cite{mythosofmodelinterpretability}.
While the meaning of a \emph{reasonable size} is ill-defined and context-specific, it indicates that model simplicity is a crucial property to consider while building these models.

\subsection{Privacy}
\label{subsec:privacy}

The development of privacy-preserving mechanisms for ML has been widely motivated by the flourishing literature on inference attacks against models in recent years. 
In the generic setting, such attacks leverage the outputs of a computation to retrieve information regarding its inputs~\cite{doi:10.1146/annurev-statistics-060116-054123}.
More specifically in ML, the computation being performed is usually a learning algorithm whose output is a trained model. 
Two distinct adversarial settings are generally considered in the literature. 
In the \emph{black-box} setting, the adversary does not know the model's parameters and can only query it through an API. 
In contrast, in the \emph{white-box} setting, the adversary has full knowledge of the model parameters. 
Of course between these two extreme scenarios, diverse \emph{gray-box} settings are possible. 

Different types of inference attacks have been proposed against ML models~\cite{DBLP:journals/corr/abs-2005-08679,DBLP:journals/corr/abs-2007-07646}, among which:
\begin{itemize}
    \item \emph{{Membership inference attacks}} try to infer whether given examples were used to train a model or not~\cite{DBLP:conf/sp/ShokriSSS17}.
    \item \emph{{Reconstruction attacks}} aim at reconstructing part of a model's training data~\cite{doi:10.1146/annurev-statistics-060116-054123}. 
    \item \emph{{Model extraction attacks}} aim at stealing a black-box model's internal functionalities or parameters~\cite{DBLP:conf/uss/TramerZJRR16}. 
    \item \emph{{Model inversion attacks}} focus on retrieving a model's inputs by only observing the associated outputs~\cite{DBLP:conf/ccs/FredriksonJR15}. 
    Hence, such attacks often target the data provided at inference time (and not solely the training data).
\end{itemize}

To counter these risks, several syntactic models of anonymity were proposed. 
More precisely, these approaches 
consist in grouping examples within \emph{blocks} so that the profile of a user is indistinguishable among those belonging to the same block~\cite{DBLP:journals/tdp/CliftonT13}. 
For instance, $k$-anonymity~\cite{DBLP:journals/ijufks/Sweene02,DBLP:journals/tkde/Samarati01}, requires that each block contains at least $k$ examples. 
Several extensions of $k$-anonymity were proposed, among which $t$-closeness~\cite{DBLP:conf/icde/LiLV07} additionally ensures that the distribution of the values within each block is sufficiently close to that of the entire dataset.

Nonetheless they are not well-adapted to ML and do not provide formal privacy guarantees.
Thus, \emph{differential privacy} (DP) has been adopted as the leading approach, in parts because it can be used to precisely bound the amount of information the output of a computation leaks regarding its inputs~\cite{DBLP:conf/tcc/DworkMNS06}. 
Referring to ($\epsilon{}$,$\delta{}$)-DP, two parameters help control the level of enforced privacy. 
Intuitively, $\epsilon$ bounds the contribution of each individual example to the output of the computation, while $\delta$ corresponds to the probability of privacy failure, with tighter values of these parameters indicating a stronger privacy protection.
\emph{Pure DP} refers to scenarios in which $\delta=0$ while \emph{approximate DP} covers cases in which $\delta > 0$. 
DP exhibits several important properties, among which the immunity to post-processing, which states that the output of a differentially-private algorithm remains differentially-private whatever (data-independent) computations are further performed on it. 
Several mechanisms were proposed to enforce DP~\cite{dwork2014algorithmic}. 
For instance, the \emph{Laplace} (respectively, \emph{Gaussian}) \emph{mechanism}~\cite{DBLP:conf/tcc/DworkMNS06} adds random noise drawn from a Laplace (respectively, Gaussian) distribution to the computed value, with the noise magnitude being scaled to the function's sensitivity (\emph{i.e.,} the maximum impact a single individual can have on the computation's output).
The \emph{{functional mechanism}}~\cite{DBLP:journals/pvldb/ZhangZXYW12} approximates the function using its polynomial Taylor expansion and perturbs the coefficients of the resulting polynomial form with noise.
Unlike the aforementioned noise addition techniques, the \emph{exponential mechanism}~\cite{DBLP:conf/focs/McSherryT07} consists in drawing an output from a probability distribution, in which the probability of a candidate depends on its utility.
Several frameworks for differentially-private ML exist~\cite{ji2014differential,DBLP:journals/cim/GongXPFQ20}.
For instance, \dpsgd{}~\cite{DBLP:conf/ccs/AbadiCGMMT016} was proposed to train deep learning models under DP. 
The authors have modified the traditional Stochastic Gradient Descent (SGD) by clipping the norm of the computed individual gradients (to bound each example's contribution to the computation) before perturbating them with Gaussian noise.
Another approach based on ensemble methods, called \pate{}, considers a particular setup, with a private training set and a public unlabeled one~\cite{DBLP:conf/iclr/PapernotAEGT17,DBLP:conf/iclr/PapernotSMRTE18}.
First, the (private) training set is partitioned into a number of non-overlapping subsets used to train a set of \emph{teacher} models. 
Afterwards, the predictions of the teachers (\emph{i.e.}, vote histograms) are made differentially-private by adding Laplace noise.
The public data is then labeled using these noisy predictions, and used to train a differentially-private \emph{student} model.

%% file: 2-fairness-interpretability.tex
In this section, we first review the tensions between fairness and interpretability before exploring some synergies.

\subsection{Tensions}
\label{subsec:sota_fairness_interpretability_tensions}

First, we elaborate on the theoretical and empirical tensions between fairness and simplicity, which is often considered as a proxy for interpretability.
Afterwards, we discuss the main challenges that need to be tackled when jointly pursuing the interpretability and fairness desiderata.
Finally, we list different ways in which post-hoc explanations can be unfair.

\subsubsection{Tensions between Fairness and Simplicity}
\label{subsec:sota_fairness_interpretability_tensions_1}

\paragraph*{\textbf{Simplicity and fairness intrinsically conflict}} 
A framework to theoretically study the implications of enforcing interpretability is proposed in~\cite{agarwalfairnessinterprworkshop}, adapted from that of~\cite{kleinberg2019simplicity}. 
It considers simplicity as a proxy for interpretability.
More precisely, a ML model is represented as a set of cells partitioning the input space and simplifying a model consists in merging some of its cells (hence diminishing their number and the model's complexity).
The authors prove that, for every non-trivial group-agnostic simplification, there exists a more complex classifier that simultaneously strictly improves both accuracy and (statistical) fairness.
This classifier can be efficiently constructed by carefully selecting some examples from chosen subgroups and splitting their associated cells.
Overall, this result suggests that interpretability/simplicity comes at some cost in terms of accuracy/fairness. 
Similar results were originally shown in~\cite{kleinberg2019simplicity}, further demonstrating how simplicity is \emph{fundamentally inconsistent} with statistical fairness notions.
As stated in~\cite{interpretabilitytradeoffs}, while model interpretability is an abstract notion, enforcing it can only reduce the set of admissible ML models. 
Consequently, ensuring interpretability can only decrease the (training) accuracy. 
A similar reasoning can also be done with respect to fairness.
More precisely, by limiting the space of admissible classifiers, the enforcement of fairness reduces the number of possible trade-offs, which can be an obstacle to achieve both fair and accurate learning. 

\paragraph*{\textbf{Empirical trade-offs are complex}}
An empirical study of the trade-offs between interpretability and fairness was conducted in~\cite{jabbari2020empirical}. 
In this study, the number of features available to a classifier is used as a measure of its complexity and acts as a proxy for interpretability.
By changing this number, the authors report the variations obtained with respect to statistical fairness notions (namely, statistical parity and equal opportunity).
Experiments on synthetic and real-world datasets show several trends, that mainly depend on the correlation between sensitive attributes, non-sensitive ones as well as class labels.
As expected, when the sensitive attribute is correlated (even moderately) with the class label, using it explicitly greatly increases the model's unfairness.
The results obtained rely strongly on the chosen notion of interpretability and as such cannot be considered generic.
In addition, they demonstrate that the trade-off between fairness and interpretability is, in practice, complex and data-dependent.

\subsubsection{Combining Fairness and Interpretability is Challenging}
\label{subsec:sota_fairness_interpretability_tensions_2}

\paragraph*{\textbf{Learning optimal interpretable models under fairness constraints is computationally challenging}}  
Due to their combinatorial nature, learning optimal interpretable machine learning models under constraints (\emph{e.g.}, fairness constraints) has been identified as one of the main technical challenges towards interpretable machine learning~\cite{rudin2022interpretable}. 
While approaches producing optimal interpretable and fair ML models exist in the literature (\emph{e.g.,} an Integer Programming formulation for learning optimal fair decision trees), they are often computationally expensive and difficultly scale.
Yet, recent work shows that the conflict between accuracy and fairness can be leveraged to perform an effective pruning (using Integer Linear Programming) when learning optimal fair rule lists~\cite{DBLP:conf/cpaior/AivodjiFGHS22}.

\paragraph*{\textbf{Explanations may not preserve fairness properties of a model}} 
It was observed in~\cite{DBLP:journals/corr/abs-2106-13346} that popular explainability frameworks may not reliably reflect the fairness properties of the explained models. 
For example, on the one hand it is possible to compute post-hoc explanations that appear to be fair to explain an unfair black-box model~\cite{fairwashing}.
On the other hand, the explanations of a fair model's decisions may (wrongly) rely on sensitive features and exhibit discrimination~\cite{DBLP:conf/aies/ManerbaG22}.
In addition, the choice of the explanation method as well as the type of explanation it produces both impact the users' perceived fairness~\cite{dodge2019explaining}.
The fairness of post-hoc explanations generated from a fair model's decisions was also investigated in~\cite{DBLP:journals/corr/abs-2106-13346}. 
More precisely, based on group fairness notions, the fairness of an explanation can be formulated similarly to that of a classifier (an explanation being seen as a local surrogate model). 
Afterwards, fairness is computed on a neighbourhood of the explained example. 
For such artificial points, no label is known, which means that only the statistical parity metric can be used.
These researchers show that the fairness property of the explained model may not be reflected in the generated explanations and propose a framework for producing fairness-preserving explanations.

\paragraph*{\textbf{Fairness-enhancing methods may require non-interpretable transformations, hence harming interpretability}} 
In a study on interpretable, fair and accurate ML for criminal recidivism prediction, \cite{wang2022pursuit} observe that fairness-enhancing methods often require non-interpretable transformations, which are not compatible with interpretability desiderata.
Indeed, pre-processing methods usually perform complex transformations of the input features, which harm their original semantic~\cite{kamiran2012data,Zemel2013}. 
The resulting representation hence can not be used to produce an understandable model.
Furthermore, the corrections performed to a model's outputs by post-processing techniques~\cite{pleiss2017fairness} can also lead to non-interpretable processes.

\subsubsection{Other Unfair Effects of Explainability Methods}
\label{subsec:sota_fairness_interpretability_tensions_3}

\paragraph*{\textbf{Post-hoc explanations affect individuals' privacy in a disparate manner}} 
As discussed later in Section~\ref{subsubsec:fairness_privacy_tensions}, minority groups often suffer from increased privacy risks.
Interpretability can also exhibit this trend, as noted by \cite{shokri2020exploiting,shokri2021privacy}. 
For instance, when investigating whether membership information 
can be inferred from post-hoc explanations, it has been observed that outliers as well as ``hard to generalize'' examples belonging to minority groups are at a higher risk of being disclosed than majority groups. 
This is partly due to the fact that they are more susceptible of being part of the generated explanations. 
In such case, interpretability tools can penalize minorities by leaking more information about disadvantaged groups. 

\paragraph*{\textbf{Post-hoc explanation frameworks can introduce unfairness through disparity in explanation quality}} Group-based disparities in explanation quality have been recently investigated in~\cite{DBLP:conf/aies/DaiUABL22}. 
More precisely, the authors first identify key characteristics that define the quality of an explanation (\emph{e.g.}, fidelity, stability, consistency and sparsity). 
Then, they conduct a large experimental study demonstrating that there is often a disparity in the quality of the explanations produced affecting minority groups. 
Such quantitative disparity is identified to depend on the type of model being explained and on the particular post-hoc explanation framework considered. 
Using several real-world applications (\emph{e.g.}, finance, healthcare, college admissions and the US justice system) and post-hoc explanation frameworks, \cite{DBLP:conf/fat/BalagopalanZHHR22} have also demonstrated that the fidelity of the produced explanations varies significantly across the different identified subgroups of the population. 
Finally, they suggest that robustness techniques can help reduce the observed disparity - but emphasize that communicating details regarding such disparity to end-users is critical.

\paragraph*{\textbf{Counterfactual explanation frameworks can harm subgroups of the population by consistently providing higher-cost recourse}} In the context of counterfactual explanations, the \emph{cost of recourse} is defined as the amount of effort a user has to do to implement the provided recourse and change the model's decisions. 
In this context, it was shown that counterfactual explanation frameworks may provide lower-cost recourse for some subgroups of the population while harming some others~\cite{DBLP:conf/fat/UstunSL19,DBLP:conf/aies/SharmaHG20}. 
For instance, some minority groups may have to make more effort to implement the provided recourse after a loan refusal. 
To face this issue, \emph{recourse fairness} was studied~\cite{DBLP:journals/corr/abs-1909-03166,DBLP:journals/corr/abs-2010-04050} and frameworks equalizing the cost of recourse across subgroups were proposed.

\paragraph*{\textbf{Post-hoc explanations can be manipulated}}
Explainability tools are designed to facilitate model audit and enhance the users' understanding. 
However, because the process of explanation generation can sometimes be opaque, post-hoc explanations could potentially be manipulated by black-box model holder to hide unfair decision-making processes by providing manipulated fair explanations. 
Indeed, it was shown that black-box explanations can be misleading, for instance by achieving high fidelity with respect to the explained model while using entirely different features, leveraging correlations in the feature space~\cite{10.1145/3375627.3375833}. 
In addition, it has been demonstrated that this can be exploited and extended to an existing framework~\cite{DBLP:conf/aies/LakkarajuKCL19} to generate explanations favoring some given features while avoiding others. 
Finally, the authors have conducted a user study and find out that misleading explanations can increase the user trust in black-box models wrongly.

Other works have also shown how malicious entities can manipulate explainability techniques to hide the true reasoning of the underlying model. 
For example, it is possible to directly craft manipulated explanations, such as local surrogate models~\cite{fairwashing,DBLP:conf/nips/AivodjiAGH21} that appear fair but actually explain the output of a globally unfair black-box, with such practice being coined as ``fairwashing''.
Explanation frameworks can also be potentially manipulated, for instance by detecting artificial examples generated by perturbation-based methods and giving them a chosen output value~\cite{foolingLIME}.
This can be leveraged to hide a black-box model's unfairness by crafting and providing fair explanations to an auditor~\cite{DBLP:journals/corr/abs-2106-12563}.
Furthermore,~\cite{DBLP:conf/nips/HeoJM19,dimanov2020you} have shown that it is possible to fine-tune a pre-trained model to manipulate the output of feature importance explanation methods while having little impact on the model's accuracy.
Considering sequence classification and sequence-to-sequence tasks (\emph{i.e.}, in which the input to the model is a sequence of words),~\cite{DBLP:conf/acl/PruthiGDNL20} propose a method to train a model with significantly reduced attention mass over some chosen words (\emph{e.g.}, gender-related prefixes) while still using them for prediction.
A user study shows that the proposed method is able to mislead users into thinking that the underlying model is fair, while it is actually biased against gender.

It was also shown to be possible to learn a model so that the counterfactual explanations generated by some off-the-shelf algorithm look \emph{recourse fair} across subgroups of the population (\emph{i.e.}, the cost of the recourse associated to the counterfactual explanations does not vary too much between individuals from the different subgroups), while also being able to generate lower-cost recourse explanations for some privileged subgroup(s) by simply adding a small adversarial perturbation~\cite{DBLP:journals/corr/abs-2106-12563,DBLP:conf/nips/SlackHLS21}.
In \cite{DBLP:conf/uss/ZhangWSJL020}, an adversary is able to generate adversarial examples with chosen prediction by a black-box model that also fool popular explainability tools. 
This illustrates the fact that post-hoc explainability techniques are not a reliable way to detect adversarial inputs manipulation.
Finally,~\cite{laberge2022fooling} consider the setup of a fairness audit in which the data is private and owned solely by the malicious model holder, which provides subsamples to the external auditor.
They show that the former can manipulate the auditor's explainability methods to hide unfair decision-making (such as the influence of a sensitive attribute) by providing adversarially-selected data samples.
In addition, such practices are particularly difficult to detect in a remote setting, in which the explanation is provided by a third-party API~\cite{bouncerproblem}.

Finally, although many tensions between explainability/interpretability and fairness exist, one can still identify some synergies, as discussed hereafter.

\subsection{Synergies}
\label{subsec:sota_fairness_interpretability_synergies}

\paragraph*{\textbf{Interpretability and explainability ease model audit}} 
As mentioned in~\cite{stopexplainingblackboxmodels}, it is easier to detect and debate possible biases or unfairness issues with an interpretable model than with a black-box one. 
This inherent benefit of interpretable models applies both to fairness and accuracy, as it makes it possible to detect and correct possible inaccuracies with respect to the training data - which is more difficult with black-box models. 
Following the same line of research,~\cite{doshi2017towards} state that interpretability can be used to qualitatively ascertain whether other desiderata - such as fairness - are met.
Post-hoc explainability methods can also facilitate fairness audit by gaining insight regarding the causes of a model's unfairness.
For instance,~\cite{begley2020explainability} propose to rely on \emph{fairness explanations} based on Shapley values to be able to attribute a model’s overall unfairness to individual input features. 

\paragraph*{\textbf{Fairness can act as a regularizer}} It was observed in the literature that enforcing fairness constraints can have a regularizing effect, thus also reducing overfitting~\cite{kilbertus2018blind}. 
More precisely by preventing over-complex models, this can lead to sparser and more interpretable models.

%% file: 3-fairness_privacy.tex
In this section, we first highlight the identified theoretical and empirical tensions between fairness and privacy. 
We then review some synergies illustrating how the two requirements can be conciliated.
Note that part of this intersection is covered in much more details by a recent survey~\cite{DBLP:conf/ijcai/Fioretto0HZ22} studying the interactions between fairness and differential privacy (DP), in both decision making and machine learning tasks.

\subsection{Tensions}
\label{subsubsec:fairness_privacy_tensions}

As discussed in Section~\ref{subsec:fairness}, it is desirable and often legally required to ensure that sensitive attributes do not directly or indirectly influence the predictions of a ML model.
However, while many popular fairness-enhancing approaches require the availability of such sensitive attributes, their collection and use may be prohibited by privacy regulations or anti-discrimination laws.
Some approaches propose to use an encrypted version of the sensitive attributes so that the users do not have to explicitly reveal this information. 
For instance,~\cite{kilbertus2018blind} leverage cryptographic approaches such as Secure Multi-Party Computation (SMPC) to build a fair model. 
Nevertheless, 
processing encrypted information ensures that the computation does not leak anything more than its outputs, but does not protect them
from inference attacks.
This illustrates a first, straightforward intrinsic conflict between fairness and privacy. 
Furthermore, when applied jointly, both notions often conflict, as discussed in more details in the following paragraphs.

\paragraph*{\textbf{Group fairness and differential privacy are theoretically incompatible}} 
It is provably impossible to build ML models strictly respecting a given group fairness constraint while respecting DP. 
More precisely,~\cite{10.1145/3314183.3323847} have shown that $(\epsilon{},0)$-DP and fairness (more precisely equal opportunity) cannot be simultaneously satisfied without reaching trivial accuracy. 
The authors have noted that this holds for pure $(\epsilon{}, 0)$-DP, but is also applicable for ($\epsilon{}$,$\delta{}$)-DP (as $\delta{}$ is usually required to be cryptographically small). 
An impossibility theorem is also stated in~\cite{1548832}, considering popular group fairness definitions: 
\emph{if a learning algorithm $\learningalgo{}$ is $(\epsilon{},0)$-differentially private and is guaranteed to output an approximately fair classifier, then $\learningalgo{}$ is constrained to output a constant classifier.}
The idea of their proof is essentially the same as~\cite{10.1145/3314183.3323847}. 
(i) Consider a learning algorithm $\learningalgo{}$ that is $(\epsilon{}, 0)$-DP. For any two datasets $\dataset{}$ and $\dataset{}'$, and for any classifier $\classifier{}$, if $\learningalgo{}$ outputs $\classifier{}$ for $\dataset{}$ with probability strictly greater than zero, then it must output $\classifier{}$ for $\dataset{}'$ with strictly positive probability too. 
This can be proved because, for any two datasets $\dataset{}$ and $\dataset{}'$, it is possible to build a serie of datasets neighboring two-by-two, from $\dataset{}$ to $\dataset{}'$ (and the property must be verified for all pairs of neighbouring datasets by definition of pure DP). 
(ii) Recall that $\learningalgo{}$ can only output classifiers respecting a given (exact or approximate) fairness requirement: if a classifier $\classifier$ does not meeet the fairness requirement on the training set $\dataset$, then $P(\learningalgo(\dataset)=\classifier)=0$. 
The conjunction of (i) and (ii) implies that $\learningalgo{}$ can only release constant classifiers (and hence pure DP and group fairness cannot be satisfied jointly). 

\paragraph*{\textbf{Enforcing fairness increases privacy vulnerabilities}} 
Disparities with respect to the vulnerability to Membership Inference Attacks (MIAs) between various subgroups of the population are observed in~\cite{kulynych2019disparate}.  
The theoretical analysis suggests that vulnerability to MIA is caused by \emph{distributional overfitting}, which quantifies the distance between the distributions of outputs of the model on the training set and outside. 
Disparate vulnerability to MIAs arises if and only if distributional overfitting differs across subgroups.
In practice, as aforementioned in Section~\ref{subsec:sota_fairness_interpretability_tensions_3}, subgroups that are inherently more difficult to fit and/or that are less represented in the data are indeed more vulnerable to MIAs.
Additionally, overfitting can increase these vulnerabilities, but also their disparities. 
For instance, it was empirically shown that enforcing fairness constraints may help under certain conditions, but can also exacerbate the observed disparities or even create new ones in real-world applications.
Finally, the authors have recalled that DP upper-bounds the vulnerability of all individuals or subgroups, hence also upper-bound their disparity.
However, it does not remove it completely and in addition to get an interesting mitigation, the privacy budget must often be really tight, hence resulting in utility drops.

In a position paper,~\cite{ekstrand2018privacy} emphasize the importance for a privacy-preserving mechanism to protect individuals with equivalent effectiveness. 
However, while DP provides the same (worst-case) theoretical protection for all dataset examples, the actual privacy vulnerability is often not uniformly distributed. 
The privacy implications of fairness are empirically studied in~\cite{DBLP:conf/eurosp/ChangS21}, quantifying the \emph{data privacy risk} as the success of a black-box MIA.
The authors have empirically shown that enforcing fairness constraints disproportionately raises the privacy risk of the unprivileged subgroups: ``fairness comes at the cost of privacy, and the privacy cost is not equal across subgroups''. 
This is explained by the fact that the fairness requirements they have used requires the model to equally fit the unprivileged subgroups. 
When such subgroups are smaller, each example has a stronger impact over the resulting model and, in the worst case, is memorized.
In addition, the more unfair the unconstrained model is, the higher the privacy vulnerability disparity will be, as there is more unfairness to be compensated.
Finally, information regarding a model's fairness can be exploited to reconstruct the sensitive attributes of its training examples~\cite{hu2020inference,ferry2023exploiting}. 
These works rely declarative programming approaches to encode the fairness desiderata and perform (or improve) the reconstruction.

\paragraph*{\textbf{Differential privacy disproportionately affects utility}} The effects of enforcing differential privacy on a model's accuracy on different subgroups of the population are studied in~\cite{DBLP:conf/nips/BagdasaryanPS19}, using the \emph{accuracy parity} fairness notion, which equalizes the model's accuracy across the subgroups.
Considering several image classification and natural language tasks, they use the popular \dpsgd{}~\cite{DBLP:conf/ccs/AbadiCGMMT016} framework for differentially-private deep learning in both centralized and federated settings. 
This large empirical study shows that gradient clipping and random noise addition, the key mechanisms of \dpsgd{}, disproportionately affect underrepresented subgroups.
Indeed, enforcing DP leads to higher accuracy drops for minorities and discriminated groups, such as darker-skinned people in the context of facial recognition, but also at the intersections of different subgroups. 
This leads to a ``poor gets poorer effect'', in which the classes with low accuracy in the non-DP setting suffer the largest accuracy drops when applying DP.
In a follow-up work, \cite{DBLP:journals/corr/abs-2106-12576} empirically observe that the differentially-private \pate{}~\cite{DBLP:conf/iclr/PapernotAEGT17,DBLP:conf/iclr/PapernotSMRTE18} framework (introduced in Section~\ref{subsec:privacy}) also has disparate impact on the resulting model's utility.
However, they report that \pate{} has smaller disparate impact compared to \dpsgd{} to reach similar privacy levels, and note that a sweet spot for the number of teachers exists, which minimizes the induced disparities.
The authors of~\cite{DBLP:conf/ccs/FarrandMST20} observe that the accuracy disparity caused by DP still occurs even when the data is slightly imbalanced, and for loose privacy guarantees. 
Indeed, two main factors were identified in the literature to explain this effect: properties of the training data, and characteristics of the DP mechanism, which are summarized and analyzed with more details in a recent survey~\cite{DBLP:conf/ijcai/Fioretto0HZ22}.

It was also observed in healthcare applications (x-ray images classification and mortality prediction in time series) that small groups and samples at the tail of the data distribution suffer from a larger accuracy drop compared to majority groups and typical examples~\cite{DBLP:conf/fat/SuriyakumarPGG21}.
Furthermore, the characteristics of DP learning mechanisms themselves are also directly related to the magnitude of the observed disparate impact. 
This encompasses the gradient clipping and noise addition mechanisms of \dpsgd{} (as aforementioned), as well as the size of the teacher ensemble and the confidence of the voting teachers in  \pate{}~\cite{DBLP:journals/corr/abs-2109-08630}. 
Different technical solutions to mitigate the disparate impact of DP on a model's utility were proposed. 
Indeed, it was shown that it is possible to modify \dpsgd{} to use different clipping bounds for the different identified subgroups~\cite{DBLP:conf/kdd/XuDW21}. 
Other work~\cite{9642428} performs early stopping based on a public validation set.
When using \pate{} in low voting confidence regimes, small perturbations may significantly affect the result of the voting result. 
To mitigate this phenomenon,~\cite{DBLP:journals/corr/abs-2109-08630} propose to use soft labels and report confidence scores associated with each target label, rather than reporting solely the label with the largest confidence.
While being heuristic as it does not guarantee any form of fairness, these approaches have been empirically shown to reduce the disparate impact caused by traditional DP mechanisms.

The disparate impact of DP mechanisms was also observed for decision tasks.
\cite{Pujol2020} have studied the setup in which agencies release differentially-private versions of their databases, that are then used for several allocation problems. 
The authors consider three real-life allocation problems using the differentially-private Census data: namely printing of election materials in minority languages, allocation of funds to school districts to assist disadvantaged children and apportionment of legislative representatives.
They demonstrated that the noise added by DP mechanisms leads to errors in the computed allocations compared to the true allocations (\emph{i.e.}, the allocations that would be decided without DP). 
The key point of their work is that this error affects the entities being allocated some resources in a disparate manner.
For instance, it is empirically shown that small school districts often benefit an overestimated allocation. 
On the other side, larger district may get a smaller allocation, which harms their enrolled children.
This effect was also observed in the literature with two main causes being identified~\cite{DBLP:conf/ijcai/Fioretto0HZ22}.
In a nutshell, the shape of the decision problem can disproportionately exacerbate the noise added by the DP data release if it involves non-linearities in its computation, such as thresholds for funds allocation. 
Additionally, post-processing steps can induce intrinsic biases. 
For instance, ensuring simple non-negativity constraints within the computed values can imply a positive bias.
It was also shown that DP mechanisms adding data-dependent noise are responsible for a more important disparity, due to the fact that, contrary to standard DP mechanisms (such as the Laplace mechanism), the effect of DP differs between entities. 
Finally, other aspects of privacy can also impact fairness. 
For instance, recent work~\cite{koch2023no} show that models designed to take into account potential future unlearning requests, which are request in which a user asks for the contribution of his data to be removed from the model,
disproportionately affects the utility for minority groups.

\paragraph*{\textbf{Differential privacy disproportionately affects the quality of post-hoc explanations}} Reference~\cite{DBLP:conf/sp/DattaSZ16} propose the notion of differentially-private post-hoc explanations, among which some aim at identifying proxy features that cause a \emph{group disparity} (\emph{i.e.}, a difference in the average prediction between several subgroups). 
Then, it is shown that, for minority groups, the amount of noise required to make the explanations differentially-private results in a significant loss in its utility, hence making more difficult the discovery of discriminatory proxy features.
While proposing a framework to generate differentially-private post-hoc explanations, \cite{DBLP:conf/fat/PatelSZ22} have observed that sparse data regions, which often correspond to underrepresented subgroups are associated to poorer performances, either in terms of required privacy budget or explanation quality. 
In both cases, privacy disproportionately affects minority groups, which is consistent with previously mentioned works.


Overall, DP and statistical fairness are both theoretically incompatible and strongly conflict in practice.
On the one side, to ensure fairness minority groups, the corresponding examples shall yield a higher importance in the learning process, which exposes their information more than for examples of the majority group. 
On the other side, to ensure DP, one must reduce more the influence of underrepresented subgroups, as learning an equivalent amount of information for them would result in an increased per-example privacy risk. 
Nevertheless, in the next subsection, we show that the two notions can be jointly applied under certain circumstances, and thus that there are some synergies between privacy and fairness.

\subsection{Synergies}
\label{subsec:sota_fairness_privacy_synergies}

\paragraph*{\textbf{Differential privacy and approximate 
fairness can be jointly enforced with some trade-offs}}  
As discussed in Section~\ref{subsubsec:fairness_privacy_tensions}, it is impossible for a learning algorithm to satisfy DP while also producing a model strictly complying with fairness constraints.
However, it is possible for a DP learning algorithm to output a model \emph{approximately} satisfying given fairness criteria~\cite{10.1145/3314183.3323847}. 
This leads to a trade-off between the DP guarantees and the observed model's fairness.
Hereafter, we first introduce different methods of the literature jointly handling differential privacy and fairness.

The notion of Private and Approximately Fair Agnostic PAC (Probably Approximately Correct) Learning was introduced in~\cite{10.1145/3314183.3323847}.
It states that a learning algorithm satisfies DP while returning an accurate and approximately fair classifier with high probability.
The authors implement this notion using the Exponential Mechanism, with a utility function being the sum of a model's error and unfairness. 
The sensitivity of the utility function being data-dependent, the Laplace mechanism is used to upper-bound it in a differentially-private manner.
This approach achieves the desiderata of privacy, fairness and accuracy, but the running time of the Exponential Mechanism scales linearly with the hypothesis class size, which is exponential for common hypothesis classes.
This motivates the need for an efficient algorithm conciliating these desiderata. 
To realize this, the authors have built upon a polynomial-time algorithm from the literature, producing approximately fair and accurate randomized classifiers with high probability.
In a nutshell, this algorithm formulates the fair learning problem as a two-player zero-sum game, between a Learner minimizing error while satisfying fairness constraints and an Auditor updating Lagragian multipliers to penalize the largest subgroup-wise fairness violations.
This algorithm is modified to satisfy DP by using a differentially-private subroutine to privately compute the players' best responses in each round.

Two methods are proposed in~\cite{DBLP:conf/www/XuYW19} to achieve jointly DP and fairness in logistic regression. 
\emph{Decision boundary fairness} is used as a notion of fairness that provably minimizes statistical parity violation. 
A first approach coined \pflr{} considers the fairness constraint as a penalty term to the objective function. 
DP is enforced using the functional mechanism~\cite{DBLP:journals/pvldb/ZhangZXYW12}.
More precisely, the objective function is approximated through its polynomial representation based on Taylor expansion before being perturbed by injecting Laplace noise into its polynomial coefficients. 
Minimizing the perturbed objective function leads to the computation of differentially-private model parameters.
A second approach, named \pflrs{} and based on the first one, takes advantage of the connection between ways of achieving differential privacy and fairness. 
More precisely, the authors noted that adding the fairness penalty is equivalent to shifting the value of some coefficients of the polynomial form of the objective function. 
Thus, they do not incorporate the fairness penalty term directly in the objective function and rather integrate it via mean-shifting the Laplace noise added to a subset of the coefficients. 
As such shift is dataset-dependent, a small part of the privacy budget is used to estimate it in a differentially-private manner. 
The Theoretical analysis as well as empirical evaluation show that \pflrs{}, by separating privacy budgets on objective function and fairness constraint, offers a more flexible framework to find good trade-offs among privacy, fairness, and utility. 

In a follow-up work,~\cite{Ding2020} 
extended \pflr{} by proposing to have two distinct privacy budgets in order to add Laplace noise with larger magnitude to the coefficients of the terms involving the sensitive attributes than to the others within the objective function.
They also propose a second approach using the relaxed functional mechanism to enforce approximate DP ($\epsilon{}$,$\delta{}$)-DP to improve on utility. 
It utilizes the extended Gaussian mechanism to perturb the objective, adding random Gaussian noise 
to the coefficients of the polynomial form of the objective function.
Empirical evaluation on real-world datasets confirms that the use of ($\epsilon{}$,$\delta{}$)-DP leads to an improved utility in all scenarios compared to pure DP. 
Furthermore, the use of two distinct privacy budgets can help enforcing stronger privacy guarantees while also reducing the correlations with the sensitive attribute, thus also improving fairness.

A differentially-private framework to train deep learning models that satisfy several popular group fairness notions was proposed in~\cite{DBLP:conf/aaai/0007FH21}. 
This approach considers the Lagrangian relaxation of the fairness-constrained learning problem, and leverages a Lagrangian dual approach to solve it: the fairness violation terms, weighted by Lagrangian multipliers, are directly added to the objective function. 
Then, the training procedure consists of iteratively repeating two successive steps: primal and dual. 
The primal update step optimizes the model parameters to minimize the objective function, given the current Lagrangian multipliers. 
Afterwards, the dual update step updates the value of the Lagrangian multiplier to approximate the stronger Lagrangian relaxation. 
To enforce differential privacy for sensitive attribute information, differential privacy is achieved at both steps, when computing the fairness violation terms or their gradients. 
In the primal update step, clipped and noisy gradients are used. 
The model parameters optimization is done on this noisy version of the objective function (in which only the fairness violation term, accessing subgroup membership which we want to protect, is impacted by the DP mechanism). 
A similar mechanism is done on the dual update step, in which constraint violations are clipped and perturbed with carefully calibrated Gaussian noise.
Extensive empirical evaluation shows that the fairness violation decreases as the privacy budget increases: thus enforcing DP leads to violating more fairness.
This is explained by the fact that relaxing the DP constraint allows either to perform more iterations (hence propagating more fairness violation information) or to inject less noise for a fixed number of iterations (hence propagating more accurate fairness violation information). 
Another surprising trend is that the model accuracy slightly decreases as $\epsilon{}$ increases. 
This is due to the fact that enforcing weaker DP allows the fairness constraints to have more impact on the objective function, hence penalizing more the accuracy.

Two fair learning algorithms have been adapted in~\cite{Jagielski2019} to satisfy both fairness (here in terms of equalized odds) and DP (with respect to the sensitive attributes).
They first consider the post-processing method of~\cite{HardtGoogle}. 
In a nutshell, given a pre-trained and possibly unfair classifier, the approach first computes its per-group per-ground truth prediction proportions. 
It then solves a Linear Program to compute per-group per-class prediction probabilities defining a fair randomized classifier.
To enforce $\epsilon{}$-DP in this setting, the authors simply add well-calibrated noise drawn from the Laplace distribution to the computed statistics before solving the LP with them. 
Theoretical analysis of how the introduced noise propagates to the solution of the LP leads to bounds on accuracy and fairness violation that are met with high probability. 
This quantifies a trade-off between accuracy, fairness and privacy: weaker DP guarantees lead to tighter bounds on accuracy and fairness, while stronger DP guarantees (satisfied by adding more noise) increase the bounds, and the possible loss on accuracy and fairness.
Experimental evaluation demonstrates that this simple method is able to provide interesting trade-offs even with small datasets but is expected to perform worst than the second approach on large ones.
The later builds upon an in-processing approach~\cite{AgarwalBD0W18}, which 
formulates the problem of learning a fair and accurate classifier as finding the equilibrium of a two-player min-max game. 
A Learner minimizes the objective function over the set of possible classifiers while an Auditor maximizes it by choosing the value of the multipliers penalizing fairness violations.
To enforce (approximate) ($\epsilon{}$,$\delta{}$)-DP, the authors add well-calibrated Laplace noise while computing the gradients of the Auditor, and use the exponential mechanism for the Learner's model selection. 
Similar to the first case, a stronger privacy guarantee (smaller $\epsilon{}$ and $\delta{}$) leads to weaker accuracy and fairness guarantees. 
However, a new trade-off can be controlled through the maximum norm of the multipliers: larger values lead to tighter fairness bounds but looser error bounds, and vice-versa.
For both approaches, introducing noise to achieve DP leads to a reduction in the fairness guarantees (in a similar manner as for accuracy).

\cite{mozannar2020fair} consider the setup in which the sensitive attributes are released using local DP (\emph{i.e.} a variant of DP in which each user locally randomizes his data before releasing it), and propose a two-step approach.
First, a classifier that is fair with respect to the noisy sensitive attributes is built, using a state-of-the-art in-processing fair learning algorithm~\cite{AgarwalBD0W18}.
Second, a modified version of a post-processing fairness-enhancing method~\cite{HardtGoogle} is used to ensure with high probability that the model is also fair with respect to the (unknown) original sensitive attributes.
For strong privacy regimes, this post-processing step is empirically shown to significantly decrease fairness violation.

\paragraph*{\textbf{The 
fairness cost of differential privacy can be theoretically bounded}}
Recent work theoretically shows that the impact of DP on fairness is bounded and can be computed to obtain non-trivial guarantees regarding the private model's fairness~\cite{mangold2022differential}. 
The underlying analysis relies on the fact that, just like a model's accuracy, common statistical fairness metrics are pointwise Lipschitz continuous with respect to the model parameters. 
Then, proving that the private model is sufficiently close to the optimal non-private one implies that their fairness are also close.
Interestingly, the theoretical bound tightens linearly with respect to the size of the training set: the ``loss of fairness'' due to privacy vanishes when the number of training examples increases.

\paragraph*{\textbf{Individual fairness and differential privacy are both robustness definitions}}
As introduced in Section~\ref{subsec:fairness}, individual fairness can be formulated as a Lipschitz condition: just like DP, it is a robustness definition~\cite{towardsformalfairnessinML}. 
More precisely,~\cite{Dwork2011} has observed that individual fairness constitutes a generalization of differential privacy. 
The authors draw an analogy between individuals in the setting of fairness and databases in the setting of differential privacy. 
Indeed, as also noted by~\cite{Zemel2013}, differential privacy requires that ``algorithms behave similarly on similar databases'', while individual fairness enforces that classifiers yield similar outcomes for similar instances.
This allows the use, for fairness purposes, of mechanisms designed for differential privacy. 
For instance,~\cite{Dwork2011} propose an efficient individually fair learning algorithm based on the Exponential mechanism~\cite{DBLP:conf/focs/McSherryT07}, resulting in provable loss bounds. 
In \cite{Jagielski2019}, the proposed privacy-preserving approach (ensuring DP for the sensitive attributes) can be seen as a relaxation of the strict notion of individual fairness proposed in \cite{towardsformalfairnessinML}.
Indeed, while the former enforces a ratio on the probabilities of different outcomes when a single example's sensitive attribute is modified, the latter enforces that the sensitive attribute is never used. 
Fairness through unawareness is then a strict, simple but certifiable way to ensure sensitive attribute privacy. 

\paragraph*{\textbf{Privacy and 
fairness can enhance each other in particular setups}}
The authors of~\cite{DBLP:conf/aaai/KhaliliZAS21} consider the particular setting in which a pre-trained model generates qualification scores for a set of applicants. 
These scores are then used to determine a fixed number of candidates that will be selected by the process (\emph{e.g.}, for a grant, a job\ldots).
They show that the Exponential mechanism can be used to perform the selection given the qualification scores, in order to both enforce DP for the selection process and improve fairness (here equal opportunity). 
Under some conditions regarding the properties of the subgroups, the proposed approach can make the selection procedure perfectly fair. 
Other notions of privacy can also have different interactions with fairness definitions. 
For instance,~\cite{Ruggieri} studies the context of itemset mining, in which given a dataset, the objective is to mine frequent patterns.
Then, the author shows that anonymizing the data to achieve $t$-closeness with carefully chosen parameters implies popular group fairness notions. Finally, it is possible to perform statistically significant fairness audits using differentially private sensitive attributes, taking into account the added noise~\cite{friedberg2023privacy}.

Other work~\cite{DBLP:journals/datamine/HajianDMPG15} also considers frequent patterns discovery, and propose two-step algorithms to jointly address non-discrimination (fairness) and privacy. 
More precisely, they first apply a privacy-preserving mechanism, before using data sanitization methods to enforce non-discrimination. 
Indeed, considering either $k$-anonymity or DP, they theoretically prove that the privacy guarantees are not affected by the later fairness-enhancing stage. 
On the contrary, they observe that applying privacy-preserving mechanisms on a sanitized data could alter the resulting patterns' fairness, either increasing or decreasing discrimination depending on the considered scenario (in line with the aforementioned tensions). 
Importantly, they empirically note that the utility loss incurred by jointly enforcing fairness and privacy is only marginally higher than that of enforcing privacy only. 
This result highlights a synergy between the two desiderata, in which the former privacy-enhancing step sometimes also improves fairness, overall leading to a smaller utility drop from the later discrimination sanitizing step. 
This trend is valid for both $k$-anonymity and DP, although the later leads to a higher utility cost.

%% file: 4-interpretability_privacy.tex
In this section, we first discuss some tensions between interpretability and privacy.
Although these notions inherently conflict, we then highlight synergies between them, before summarizing existing frameworks addressing them jointly. 

\subsection{Tensions}
\label{subsec:sota_interpretability_privacy_tensions}

\paragraph*{\textbf{Interpretability/Explainability and Privacy conceptually have antagonist goals}} While interpretability and privacy protection are both important requirements for responsible machine learning, they intrinsically pursue contrasting objectives~\cite{datta2023position}.
Indeed, on one hand, interpretability aims at providing more information to enhance users' understanding of a model's behavior. 
On the other hand, privacy requires a tight control of the leaked information, often obfuscating part of it to protect individuals' data. 
Jointly addressing both desiderata hence necessitates some form of arbitration~\cite{banisar2011right}.

\paragraph*{\textbf{Explainability tools can be used with the purpose of designing attacks against machine learning models}} 
Tools from explainable AI can be leveraged by malicious entities to perform more effective attacks against machine learning based systems. 
For instance,~\cite{DBLP:conf/uss/SeveriMCO21} studied malware detection models, that are usually trained on crowd-sourced data to distinguish between malicious softwares (malwares) and legitimate ones. 
The authors investigated backdoor poisoning attacks, in which an attacker injects carefully chosen datapoints to the crowd-sourced training set, resulting in its chosen malware being wrongly classified as legitimate by the detection model. 
In this context, they leverage Shapley values to identify highly effective features and their values, and efficiently craft the poisoned examples.
Explainable AI techniques were also leveraged to fool ML-based authentication systems, which 
take as input a user ID along with some fingerprinting authenticating the user uniquely.
An attacker can then use perturbation-based feature explanation techniques on a local surrogate model to efficiently craft a fingerprint authenticating a desired user given its ID~\cite{DBLP:journals/corr/abs-1810-00024}.
Again, the feature importance explanations help guiding the malicious crafting process by indicating which features most influence the decision.
A counterfactual explanation framework is modified in~\cite{DBLP:journals/tifs/KuppaL21} to generate adversarial examples. 
Counterfactual explanations of a black-box model are also used to identify the features that influence the model's decision boundaries and generate examples to conduct backdoor poisoning attacks.

\paragraph*{\textbf{Post-hoc explanations can be exploited to perform or improve inference attacks}}
Inference attacks traditionally query a model (\emph{e.g.}, via a prediction API) and use its outputs to achieve their goal, for instance determining an individual's membership in the training data, reconstructing part of the training dataset, extracting the model itself, or inferring an individual's missing attributes~\cite{doi:10.1146/annurev-statistics-060116-054123,DBLP:journals/corr/abs-2005-08679}.
Post-hoc explainability techniques, by offering explanations as additional outputs, expose a new attack surface. 
Several works showed that such explanations, whatever form they take (\emph{e.g.}, example-based, feature-based \ldots), can be leveraged to enhance the different types of privacy attacks (introduced in Section~\ref{subsec:privacy}):

\begin{itemize}[leftmargin=*]
  \item \textbf{Model extraction attacks.}  
Gradient-based (a class of feature-based) explanations of a black-box model can be exploited by an adversary to reconstruct the underlying model~\cite{milli2019model}. 
In the considered setup, the adversary owns an auxiliary dataset and can query the black-box model to obtain the model's gradients as explanations for given input points.
The authors have designed a near-optimal algorithm, which provably extracts the entire underlying model within a bounded number of queries, in the particular case in which it is a two-layer neural network with ReLU activations.
For the general case, they design an effective heuristic inspired by previous works on standard reconstruction attacks against prediction APIs.
More precisely, the attacker trains a surrogate model mimicking the black-box behavior and optimizes to match its gradients thanks to the provided explanations.
The results obtained demonstrates that model extraction from gradient explanations requires orders of magnitude less queries than from the sole predictions.
Another approach~\cite{DBLP:journals/corr/abs-2107-08909} also consider gradient-based explanations, but assume no auxiliary dataset. 
In such case, the data used to query the black-box and train the surrogate model is outputted by a generative model, which in turn tries to generate examples so that the surrogate disagrees with the black-box. 
The generative model is updated leveraging the provided gradient explanations, which dramatically reduces the required number of iterations (and queries to the black-box).
Furthermore, \cite{modelextractionfromcounterfactualexplanations} show that providing counterfactual (a class of example-based) explanations (CFs) can help to realize model extraction attacks with better precision and limited number of requests. 
More precisely, the adversary queries the black-box model with a given attack set, and trains a surrogate using the predictions of both the attack set instances and the provided CFs. 
The authors empirically show that the use of the provided CFs improves the attack by both increasing the built surrogate's fidelity with respect to the black-box model, and dramatically decreasing the required number of queries. 
A similar approach is proposed in~\cite{DBLP:journals/tifs/KuppaL21}, leveraging knowledge distillation techniques to train the surrogate model, which may mitigate the potential performance harm of an architecture mismatch between the actual black-box model and the reconstructed surrogate.
CFs provided by Machine-Learning-as-a-Service (MLaaS) platforms are also exploited in~\cite{DBLP:conf/fat/WangQM22}, which propose an efficient querying strategy to steal the underlying classification model.
Their strategy is based on the following observation: the generated CFs usually lie close to the decision boundary, while the attack set examples do not necessarily. 
This leads to a ``decision boundary shift issue'', in which the surrogate model's decision boundary is shifted compared to that of the actual black-box.
To circumvent this issue, the authors propose to generate counterfactuals for the CFs themselves, and to use them all for training the surrogate.

\item \textbf{Membership inference attacks.}
Feature-based explanations are leveraged in~\cite{shokri2021privacy} to perform MIAs.
More precisely, they consider both backpropagation-based (\emph{i.e.}, gradient-based) and perturbation-based explanations.
On one hand, they demonstrate that the former leak information regarding membership, and can effectively be leveraged to perform MIAs.
In particular, the explanations' variance is very informative, in the sense that explanations of training examples usually exhibit a low variance, while for unseen examples, this value can be considerably higher.
This is due to the fact that for training examples, the model is usually very confident, as it was optimized on them, and small perturbations are likely to not change its predictions. 
On the contrary, unseen samples can be closer to the decision boundary, which results in some features having a great impact on the model's predictions (hence high gradients norms), and the resulting explanation having high variance.
On the other hand, they further show using two popular perturbation-based frameworks~\cite{ribeiro2016should,smilkov2017smoothgrad} that the later is more resistant to membership inference.
This may be explained by the fact that perturbation-based frameworks often generate perturbed examples that lie out of the data distribution~\cite{kumar2020problems}. 
The black-box model behavior on such examples is unspecified, and so querying it with them does not provide insightful information to perform inference attacks. 
This also suggests that the resulting explanations may qualitatively by poorer: ``privacy comes at the cost of explanation quality''.
Counterfactual explanations are leveraged in~\cite{DBLP:journals/tifs/KuppaL21} to conduct MIAs.
More precisely, the black-box model is queried with an auxiliary dataset and then the model's outputs and generated counterfactual examples are used to train a shadow model.
Membership of a given example is then established by comparing the difference in prediction probabilities between the shadow model and the actual black-box to a threshold.

\item \textbf{Dataset reconstruction (and membership inference) attacks.} 
An example-based explainability framework based on influence functions~\cite{koh2017understanding} and returning influential training examples that most contribute to an example's prediction is considered in~\cite{shokri2021privacy}.
Because they explicitly reveal training points, and a training point is likely to be used to explain itself, such explanations are highly vulnerable to MIAs.
Indeed, this class of explanations allows for stronger attacks, such that dataset reconstruction attacks. 
The authors propose two algorithms that leverage the provided example-based explanations to reconstruct (part of) the model's training set.
The first algorithm is based on subspace reduction and comes with a certifiable lower bound on the number of points it discovers.
Empirical evaluation shows that it can be used to retrieve most of the training dataset for high dimensional data.
The second one is heuristic and offers no theoretical guarantees, but works well in practice for low dimensional data.
It simply consists in using previously revealed points to reveal new points. 
This naturally defines an influence graph structure over the training set, in which an edge between two training examples means that one is provided as an explanation for the other.
The proposed algorithm can then be used to explore entire Strongly Connected Components within this graph.
\item \textbf{Model inversion attacks.} 
The authors of~\cite{zhao2021exploiting} propose model inversion attacks that aim at reconstructing a black-box model's inputs given its outputs (here, its prediction along with some \emph{feature-based explanation}), hence harming the privacy of test instances\footnote{This differs from the previously mentioned reconstruction attacks. Indeed, in reconstruction attacks, the objective of the adversary is to infer information regarding the model's training data. 
In the discussed model inversion attacks, the objective is to gain information about the examples provided to the model at inference time, by only observing the model's outputs (\emph{cf.}, Section~\ref{subsec:privacy}).} (\emph{i.e.}, active users of the model). 
In the context of image-based tasks, they focus on different types of saliency map explanations to reconstruct the target model's input images, namely gradient-based explanations~\cite{simonyan2013deep}, influence-based explanations~\cite{ramaswamy2020ablation} (obtained by multiplying each input feature by its associated gradient), activation-based explanations \cite{selvaraju2017grad} and layer-wise relevance 
propagation~\cite{bach2015pixel} (\emph{i.e.}, attributing pixels' importance by backpropagating neurons' relevance).
The proposed attack uses an attack model, trained on an independent auxiliary dataset to predict images (given as input to the target model) given predictions and explanations (outputted by the target model).
As expected, the frameworks directly using the input within the explanation computation (\emph{i.e.}, influence-based ones) leak more information regarding the model's inputs, hence allowing better attack results.
Importantly, the paper shows that even non-explainable models can be attacked, leveraging attention transfer to build an explainable surrogate whose explanations are used to conduct the attack.
With a same attack objective, \cite{DBLP:conf/ccs/LuoJX22} have shown that Shapley value-based explanations provided by popular Machine Learning as a Service (MLaaS) providers can be exploited to reconstruct the private model inputs. 
They provide an information-theoretical analysis of the relationship between an example and its associated Shapley values, and demonstrate that an adversary can always infer useful information about the former using the later.
This analysis also holds for sampling-based Shapley-values, 
which are commonly computed as an efficient approximation of the exact Shapley values.
They then studied two distinct adversarial settings, 
and have shown that even an adversary with no background knowledge can reconstruct most of the private model's input examples given only its outputs and explanations.
\item \textbf{(Sensitive) attribute inference attacks.}
Sensitive attribute inference attacks can leverage feature-based model explanations, computed either with backpropagation-based or perturbation-based methods~\cite{DBLP:conf/cikm/DudduB22}.
The authors consider the two scenarios where the sensitive attribute is (or not) used for training the model and for inference. 
In both studied scenarios, the adversary leverages an auxiliary dataset to train an attack model to predict an example's sensitive attribute given only the outputs of the target model (prediction and explanation). 
They empirically show that their attack is able to leverage such explanations to perform attribute inference attack. 
Furthermore, they suggest that model explanations lead to higher attack success compared to model predictions, hence constituting a stronger attack surface to exploit.
\end{itemize}

\paragraph*{\textbf{Interpretable models inherently leak information regarding their training data}} 
The approach of~\cite{DBLP:conf/dbsec/GambsGH12} exploits the structure of a trained decision tree to reconstruct a probabilistic version of its training set. 
It is generalized in~\cite{ferry2023probabilisticd} to handle more generic types of knowledge and reconstruct probabilistic datasets from other types of interpretable models.
Both works use tools from the information theory to precisely quantify the amount of knowledge interpretable models encode, through their structure, regarding their training data.

\paragraph*{\textbf{Providing useful yet privacy-protective explanations remains an open challenge}}
As discussed in the next subsection, 
differentially-private explainability tools have been proposed, but always imply some trade-off between the explanation quality, the privacy guarantee and the model utility. 
Furthermore,~\cite{milli2019model} recall that DP can help guard against attacks from prediction APIs, but it is not clear if this is a viable approach for preventing reconstruction from explanations. 
On the same line, \cite{shokri2021privacy} state that ``the effect of DP techniques (notably the randomness they induce) on model transparency is unknown.''
Furthermore, the effect of DP on the explanations' robustness and user trust are still to be investigated~\cite{modelextractionfromcounterfactualexplanations}.

Overall, applying explainability techniques while preserving formal privacy guarantees is challenging. 
In the next subsection, we nevertheless how this could be achieved, but this implies some cost on either one aspect or the other.

\subsection{Synergies}\label{subsec:sota_interpretability_privacy_synergies}

\paragraph*{\textbf{Interpretability eases model audit and can be leveraged for privacy purposes}} Interpretability can be used to confirm other desiderata of ML systems, such as privacy~\cite{doshi2017towards}. 
It also makes it easier to detect possible privacy issues when building interpretable models~\cite{stopexplainingblackboxmodels}.
Furthermore, this auditable nature is particularly appreciated in the area of ML-based cybersecurity systems~\cite{DBLP:journals/corr/abs-2206-03585}. 
Indeed, machine learning models have shown great abilities to detect abnormal behaviors or intrusions. 
However, their black-box nature and lack of certification can be problematic as it possibly introduces weaknesses inside the security system. 
By providing an understanding of the underlying mechanisms and reasoning of the model, interpretability techniques can be helpful to detect overfitting, or in cases in which the model captures noise or inaccurate values in the data. 
This allows deploying more trustworthy models, but also helps the administrators identify potential breaches. 

\paragraph*{\textbf{Interpretability can be conciliated with privacy with some trade-offs}}
The authors of~\cite{DBLP:conf/kdd/FriedmanS10} study data mining with DP guarantees, considering decision tree learning as an illustrative task. 
They demonstrate that the design of the privacy preserving mechanism is crucial, and that there is a huge difference in terms of model utility and required sample size between a naive implementation using a general purpose privacy preserving data interface and a task-specific differentially-private learning algorithm.
Their empirical study demonstrates the ability of their proposed algorithm to learn differentially-private decision trees with reasonable cost in terms of accuracy. 
Several other works also tackled differentially private decision tree building, as summarized in~\cite{DBLP:journals/csur/FletcherI19}.
Locally Linear Maps (LLMs) are studied in~\cite{DBLP:conf/aaai/HarderBP20} and consist in a linear combination of logistic regressions for each possible class. 
Such interpretable models are suitable to provide local explanations (using the appropriate LLM) but also global ones, as the coefficients of each class's LLMs provide insights regarding which features really matter to it.
The authors propose a procedure to learn LLMs under DP, leveraging mechanisms from the \dpsgd{} framework~\cite{DBLP:conf/ccs/AbadiCGMMT016}. 
They empirically observe a trade-off between the privacy guarantee and the model's accuracy and interpretability.

\paragraph*{\textbf{Post-hoc Explainability can be conciliated with privacy with some trade-offs}}
Quantitative Input Influence (QII) is a framework leveraging Shapley values to provide feature-based explanations quantifying the influence of input features over the model's predictions~\cite{DBLP:conf/sp/DattaSZ16}.
As such measures may leak information regarding individual users, the authors introduce a mechanism to generate differentially-private explanations to the so-called transparency queries.
Providing pure DPy guarantees, it consists in adding Laplace noise to the query answers, scaled to the query function sensitivity. 
As the proposed measures generally have low sensitivity, the amount of added noise remains reasonable which results in relatively small average utility losses. 
Nonetheless, for some types of explanations with exceptionally high sensitivity, the amount of noise added may significantly harm their utility.
A method to generate differentially-private feature-based explanations (\emph{i.e.}, local linear surrogates) of a black-box model is introduced in~\cite{DBLP:conf/fat/PatelSZ22}. 
In their framework, the explanations are computed using a differentially-private gradient descent leveraging the Gaussian mechanism.
They further proposed an adaptive mechanism, reducing the spending of the privacy budget by leveraging the explanations to previous queries when computing a new one. 
Using tabular, text and image data, they empirically observe that the explanations' quality degrade while the privacy guarantees tighten.
\cite{DBLP:journals/corr/abs-2106-13203} investigated the impact of a model's differential privacy on the quality of post-hoc explanations (saliency maps~\cite{selvaraju2017grad}) of this model and on its utility, considering either local DP (classical learning algorithm applied on DP data) or global DP (differentially-private training algorithm). 
In both cases, the explanations are also differentially-private due to the post-processing property (\emph{cf.} Section~\ref{subsec:privacy}).
Handling either general or medical imaging applications, they have learnt neural networks under different DP budgets and evaluate the quality of post-hoc explanations of their predictions using two metrics from the literature. 
In a nutshell, these metrics aim at quantifying how much the regions highlighted by explanation maps actually account for the explained decisions.
The experimental results show that these metrics degrade while the privacy budget is tightened.
Furthermore, they suggest the existence of a three dimensional trade-off space between privacy, explanation quality and model accuracy.
To face the explanation-guided backdoor poisoning attack studied in~\cite{DBLP:conf/uss/SeveriMCO21} (and discussed in Section~\ref{subsec:sota_interpretability_privacy_tensions}),~\cite{nguyen2022xrand} proposed to generate Locally Differentially Private explanations. 
By randomly perturbating the top-$k$ features in the generated feature-based explanations, the mechanism is shown to mitigate the success of the attack. 
An approach to generate robust counterfactual explanations for differentially private Support Vector Machines (SVMs) is designed in~\cite{DBLP:journals/corr/abs-2102-03785}. 
More precisely, privacy is achieved by adding Laplace noise to the SVMs' weights, and classical counterfactual explanation frameworks may generate counterfactuals that allow to cross the classifier's noisy boundaries, but not to actually change the example's class in real-life. 
To address this issue, they instead generate robust counterfactual explanations by solving an optimization problem with probabilistic constraints. 
In practice, the generated counterfactuals require more and more changes to the example as the privacy level tightens, in order to ensure that its classification changes with respect to the (unknown) non-private classifier. 
Again, this illustrates the trade-off between explanations quality and privacy protection.
In the context of federated learning, \cite{DBLP:journals/corr/abs-2302-08044} have also noticed that DP can alter the meaningfulness of gradient-based explanations. 
They propose an adaptive mechanism still providing DP guarantees but injecting noise within the model's parameters in a manner aimed at preserving the quality of gradient-based explanations. 
Finally, recent work also studied DP for counterfactual explanations~\cite{DBLP:journals/corr/abs-2208-02878}. 
The approach consists in using an autoencoder trained in a differentially-private manner to build noisy class prototypes, which can then be leveraged to generate the counterfactuals.

\bigskip

%% file: 5-conclusion.tex
We have seen throughout this paper that while fairness, interpretability and privacy are three important dimensions of responsible ML, they often conflict in different ways, both theoretically and empirically. 
Nonetheless, we have also identified synergies, which suggests that 
a careful design can sometimes lead to improving them jointly with a reduced impact on utility.
However, this considerably increases the complexity of the learning process while requiring an in-depth analysis of the used techniques.
Furthermore, compromises usually have to made.
Overall, learning a model with non-trivial utility and satisfying our three desiderata  requires a thorough theoretical formulation, being aware of the existing tensions as well as of common techniques to mitigate them.
\revision{Both are summarized in Figures~\ref{fig:summary_figure_compatibilities_synergies} and~\ref{fig:summary_figure_tensions}, in the Appendix~\ref{appendix:summary}.}



Finally, it is crucial to promote an interdisciplinary approach, for computer scientists to ensure that the metrics they optimize for actually match legal and ethical requirements. 
This is a particularly challenging aspect: ethical analysis are often strongly context-dependent while genericity is a common objective in computer science.
In addition, not all legal and ethical notions can easily be implemented and quantified using mathematical formulas. 
It is hence necessary to verify the alignment of the notions we use with the concepts we target, for the development of ML systems that can be trusted and that do not harm the society.

%% file: 999-appendix_summary.tex
\section{Summary Figures}
\label{appendix:summary}

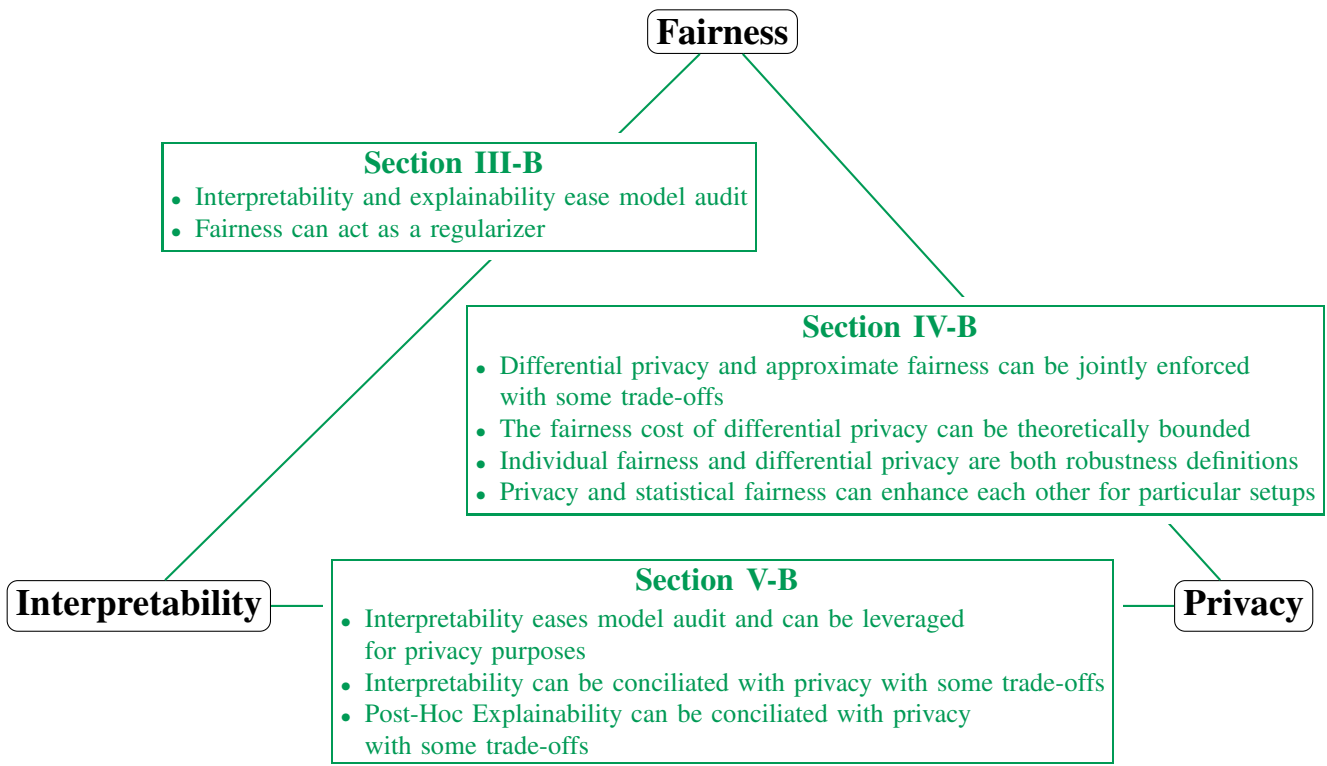
\begin{figure*}[b]
    \begin{center}
    \begin{tikzpicture}
        \node[main node] (1) {\Large \textbf{Fairness}};
        \node[main node] (2) [below left = 7cm and 5cm of 1]  {\Large \textbf{Interpretability}};
        \node[main node] (3) [below right = 7cm and 5cm of 1] {\Large \textbf{Privacy}};
    
        \path[draw,thick]
        (1) edge[ForestGreen] node[pos=0.45, yshift=1.25cm, fill=white] {\framebox{{\begin{varwidth}{\linewidth}
        {\hfill \large \textbf{Section~\ref{subsec:sota_fairness_interpretability_synergies}} \hfill}
        \begin{itemize}[leftmargin=*]
        \item Interpretability and explainability ease model audit 
        \item Fairness can act as a regularizer
        \end{itemize}
        \end{varwidth}}}} (2) 
        (2) edge[ForestGreen] node[midway, yshift=0.75cm, below, fill=white] {\framebox{{\begin{varwidth}{\linewidth}
        {\hfill \large \textbf{Section~\ref{subsec:sota_interpretability_privacy_synergies}} \hfill}
        \begin{itemize}[leftmargin=*]
        \item Interpretability eases model audit and can be leveraged\\ for privacy purposes
        \item Interpretability can be conciliated with privacy with some trade-offs
        \item Post-Hoc Explainability can be conciliated with privacy\\ with some trade-offs
        \end{itemize}
        \end{varwidth}}}} (3)
        (3) edge[ForestGreen] node[pos=0.68, yshift=-2.5cm, fill=white] {\framebox{{\begin{varwidth}{\linewidth}
        {\hfill \large \textbf{Section~\ref{subsec:sota_fairness_privacy_synergies}} \hfill}
        \begin{itemize}[leftmargin=*]
        \item Differential privacy and approximate fairness can be jointly enforced\\ with some trade-offs
        \item The fairness cost of differential privacy can be theoretically bounded
        \item Individual fairness and differential privacy are both robustness definitions
        \item Privacy and statistical fairness can enhance each other for particular setups
        \end{itemize}
        \end{varwidth}}}} (1);
    \end{tikzpicture}
    \end{center}

    \caption{Summary of the identified compatibilities and synergies between fairness, interpretability and privacy in machine learning.}
    \label{fig:summary_figure_compatibilities_synergies}
\end{figure*}

\begin{figure*}    
    \begin{center}
    \begin{tikzpicture}
        \node[main node] (1) {\Large \textbf{Fairness}};
        \node[main node] (2) [below left = 12cm and 5cm of 1]  {\Large \textbf{Interpretability}};
        \node[main node] (3) [below right = 12cm and 5cm of 1] {\Large \textbf{Privacy}};
    
        \path[draw,thick]
        (1) edge[red] node[pos=0.60, yshift=2cm, fill=white] {\framebox{{\begin{varwidth}{\linewidth}
        {\hfill \large \textbf{Section~\ref{subsec:sota_fairness_interpretability_tensions}} \hfill} \break
        \textbf{\emph{Tensions between Fairness and Simplicity - Section~\ref{subsec:sota_fairness_interpretability_tensions_1}}}
        \begin{itemize}[leftmargin=*]
        \item Simplicity and fairness intrinsically conflict
        \item Empirical trade-offs are complex
        \end{itemize}
        \textbf{\emph{Combining Fairness and Interpretability is Challenging - Section~\ref{subsec:sota_fairness_interpretability_tensions_2}}}
        \begin{itemize}[leftmargin=*]
        \item Learning optimal interpretable models under fairness\\ constraints is computationally challenging
        \item Explanations may not preserve fairness properties of a model
        \item Fairness-enhancing methods may require non-interpretable\\ transformations, hence harming interpretability
        \end{itemize}
        \textbf{\emph{Other Unfair Effects of Explainability Methods - Section~\ref{subsec:sota_fairness_interpretability_tensions_3}}}
        \begin{itemize}[leftmargin=*]
        \item Post-hoc explanations affect individuals' privacy in a disparate manner
        \item Post-hoc explanation frameworks can introduce unfairness by providing\\ lower-quality explanations to minority groups
        \item Counterfactual explanation frameworks can harm subgroups of the\\ population by consistently providing higher-cost recourse
        \item Post-hoc explanations can be manipulated
        \end{itemize}
        \end{varwidth}}}} (2) 
        (2) edge[red] node[midway, yshift=0.5cm, below, fill=white] {\framebox{{\begin{varwidth}{\linewidth}
        {\hfill \large \textbf{Section~\ref{subsec:sota_interpretability_privacy_tensions}} \hfill}
        \begin{itemize}[leftmargin=*]
        \item Interpretability/Explainability and Privacy conceptually have opposite goals
        \item Explainability tools can be used with the purpose of designing\\ attacks against machine learning models
        \item Post-Hoc explanations can be exploited to perform\\ or improve inference attacks
        \item Interpretable models inherently leak information\\ regarding their training data
        \item Providing useful yet privacy-protective explanations\\ remains an open challenge
        \end{itemize}
        \end{varwidth}}}} (3)
        (3) edge[red] node[pos=0.80, yshift=-8cm, fill=white, font=\normalsize] {\framebox{{\begin{varwidth}{\linewidth}
        {\hfill \large \textbf{Section~\ref{subsubsec:fairness_privacy_tensions}} \hfill}
        \begin{itemize}[leftmargin=*]
        \item Group fairness and differential privacy are theoretically incompatible
        \item Enforcing fairness increases privacy vulnerabilities
        \item Differential privacy disproportionately affects utility
        \item Differential privacy disproportionately affects the quality of post-hoc explanations
        \end{itemize}
        \end{varwidth}}}} (1);
    \end{tikzpicture}
    \end{center}

    \caption{Summary of the identified tensions between fairness, interpretability and privacy in machine learning.}
    \label{fig:summary_figure_tensions}
\end{figure*}
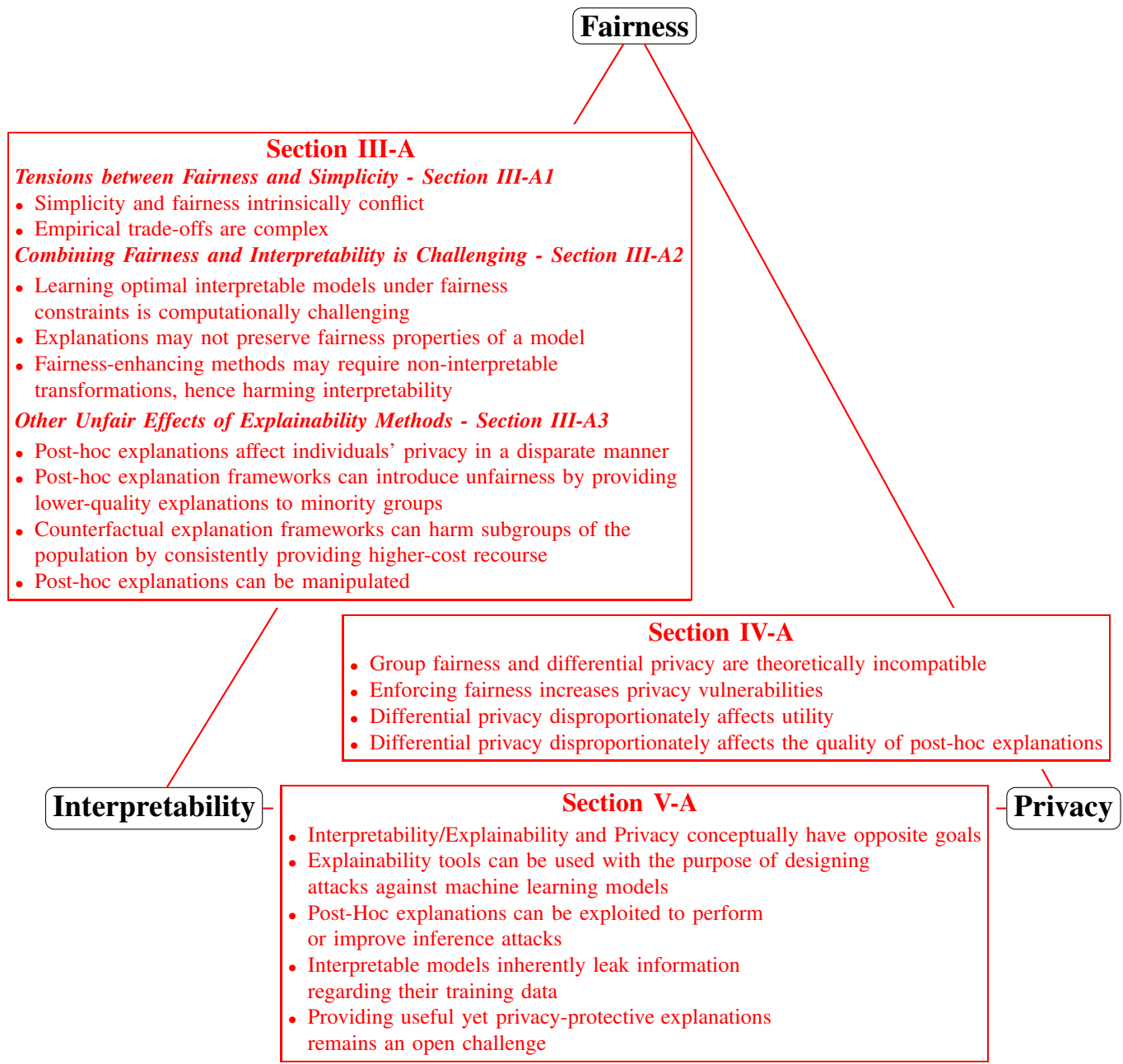

In this appendix section, we provide a graphical summary of the key interplays identified between fairness, interpretability and privacy in machine learning.
More precisely, we report compatibilities and synergies in Figure~\ref{fig:summary_figure_compatibilities_synergies}, while we overview tensions in Figure~\ref{fig:summary_figure_tensions}.

